\documentclass[letterpaper]{article} 
\usepackage{aaai2026}  
\usepackage{times}  
\usepackage{helvet}  
\usepackage{courier}  
\usepackage[hyphens]{url}  
\usepackage{graphicx} 
\usepackage{natbib}  
\usepackage{caption} 
\usepackage{algorithm}
\usepackage{algorithmic}
\usepackage{amsmath}
\usepackage{booktabs}
\usepackage{multirow}

\usepackage{newfloat}
\usepackage{listings}
\DeclareCaptionStyle{ruled}{labelfont=normalfont,labelsep=colon,strut=off} 
\floatstyle{ruled}
\newfloat{listing}{tb}{lst}{}
\floatname{listing}{Listing}
\title{Remember Me: Bridging the Long‑Range Gap in LVLMs with Three-Step Inference-Only Decay Resilience Strategies}

\author{
    Peng Gao\equalcontrib\textsuperscript{\rm 1,\rm 2},
    Yujian Lee\equalcontrib\textsuperscript{\rm 1, \rm 2},
    Xiaofeng Zhang\equalcontrib\textsuperscript{\rm 3}\thanks{Corresponding Author.},
    Zailong Chen\textsuperscript{\rm 4},
    Hui Zhang \textsuperscript{\rm 2}
}

\affiliations{
    \textsuperscript{\rm 1}Hong Kong Baptist University, HongKong, China \\
    \textsuperscript{\rm 2}Beijing Normal Kong Baptist University, ZhuHai, China \\
    \textsuperscript{\rm 3}Shanghai Jiao Tong University, ShangHai, China \\
    \textsuperscript{\rm 4}University of Wollongong, New South Wales, Australia 

}

\usepackage{bibentry}

\begin{document}

\maketitle

\begin{abstract}
Large Vision-Language Models (LVLMs) have achieved impressive performance across a wide range of multimodal tasks. However, they still face critical challenges in modeling long-range dependencies under the usage of Rotary Positional Encoding (ROPE). Although it can facilitate precise modeling of token positions, it induces progressive attention decay as token distance increases, especially with progressive attention decay over distant token pairs, which severely impairs the model's ability to remember global context. To alleviate this issue, we propose inference-only \textbf{T}hree-step \textbf{D}ecay \textbf{R}esilience \textbf{S}trategies (T-DRS), comprising (1) Semantic-Driven DRS (SD-DRS), amplifying semantically meaningful but distant signals via content-aware residuals, (2) Distance-aware Control DRS (DC-DRS), which can purify attention by smoothly modulating weights based on positional distances, suppressing noise while preserving locality, and (3) re-Reinforce Distant DRS (reRD-DRS), consolidating the remaining informative remote dependencies to maintain global coherence. Together, the T-DRS recover suppressed long-range token pairs without harming local inductive biases. Extensive experiments on Vision Question Answering (VQA) benchmarks demonstrate that T-DRS can consistently improve performance in a training-free manner. 
\end{abstract}

\begin{links}
    \link{Extended version}{https://github.com/labixiaoq-qq/Remember-me}
\end{links}

\section{Introduction}
Large Vision-Language Models (LVLMs) have demonstrated remarkable capabilities in perceiving and understanding complex multimodal information, enabling a broad spectrum of downstream tasks, referring expression comprehension, and multimodal dialogue~\cite{liu2023visual, dai2023instructblip, yang2025streamagent, zhu2023minigpt, gao2025contextual, TM-align}. These models are typically built upon large-scale pretrained language models and extended to handle visual inputs through image encoders and cross-modal alignment modules. By effectively integrating interleaved sequences of images and texts, LVLMs can provide contextually grounded and semantically rich responses to user queries. At the architectural level, most LVLMs are based on the Transformer framework~\cite{vaswani2017attention}, which has become the de facto standard for modeling sequential and structured data. However, despite its representational power, the vanilla Transformer architecture is inherently permutation-invariant and lacks a built-in notion of order. That is, its self-attention mechanism computes token-to-token dependencies without considering their positions in the sequence, making it fundamentally blind to token order. To overcome this limitation, Positional Encoding (PE) schemes are introduced to inject order-sensitive information into the input representations, enabling the model to reason about structural dependencies and temporal relations across the input sequence. 
\begin{figure}[t]
    \centering
    \includegraphics[width=0.8\linewidth]{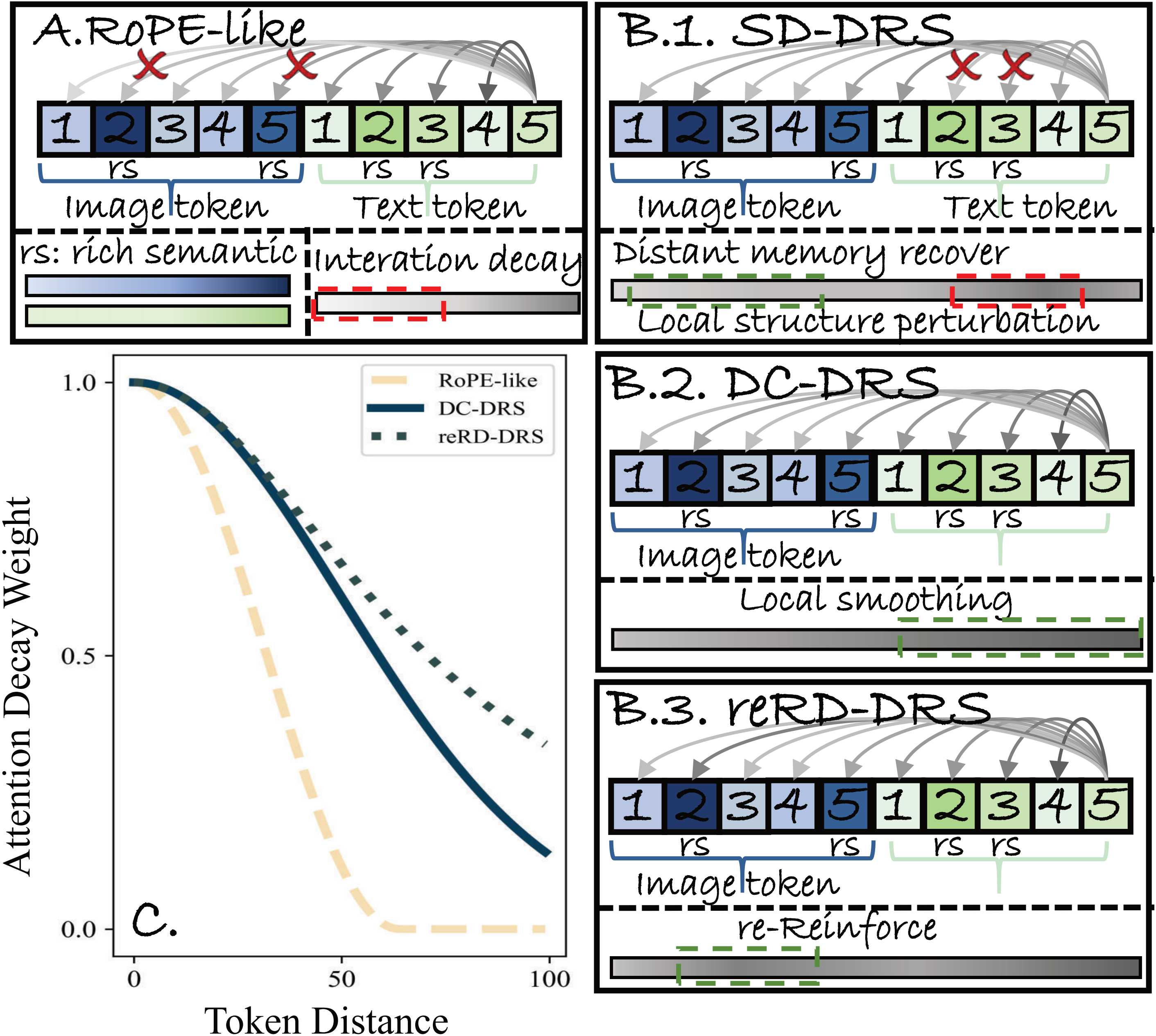}
    \caption{(A.) RoPE suffers from rapid attention decay over long token distances, weakening long-range dependencies between text and image tokens. (B.1, B.2, B.3) The proposed T-DRS framework alleviates this via three stages to collaboratively enhance and stabilize distant attention. (C.) The curves of the cross-attention weight decay in RoPE-like approaches, DC-DRS, and reRD-DRS, respectively.}
    \label{intro}
\end{figure}

PE strategies can be broadly classified into absolute and relative formulations. \textbf{Absolute PE}, which was proposed by~\cite{vaswani2017attention}, assigns each position a fixed embedding based on trigonometric functions. However, it cannot model the relative positions or distances between tokens explicitly, which limits its expressiveness, especially in scenarios requiring relational reasoning. \textbf{Relative PE}, on the other hand, captures the difference or distance between token positions, making it more suitable for encoding pairwise relations~\cite{shaw2018self, dai2019transformer}. Among relative PE variants, two approaches have gained wide adoption in modern architectures: \textbf{learnable relative embeddings} and \textbf{Rotary Position Embedding (RoPE)}~\cite{su2024roformer}. Learnable relative PE introduces trainable parameters to model positional offsets, allowing the attention mechanism to emphasize local patterns and adjacent dependencies. While effective for short sequences, such methods often exhibit poor extrapolation ability, failing to generalize beyond the maximum training length due to uninitialized or missing embeddings at test time. \textbf{RoPE}, by contrast, encodes relative positions continuously by applying complex-valued rotations. This approach maintains the distance-aware inductive bias directly within the dot-product between query and key vector, offering better generalization and compatibility with long sequences, which has been adopted in many large language and vision-language models~\cite{touvron2023llama,bai2025qwen2,liu2023visual,chen2024far}. However, as illustrated in Fig.~\ref{intro}(A), RoPE-like approaches can not be able to capture distant token information that is rich in semantics. The curve of RoPE-like in Fig.~\ref{intro}(C) exhibits a progressive decay of attention weights as the token distance increases, leading to the suppression of long-range interactions. Such decay severely limits the model’s ability to capture global context or long-range dependencies, both of which are essential in tasks requiring compositional reasoning. Several efforts explore position interpolation~\cite{chen2023extending, su2024roformer,zhu2023coca} or memory extension techniques to recover long-sequence behaviors~\cite{xing2024mitigating, zhao2025mca, li2025hope, tang2025seeing, barbero2024round} to mitigate this limitation. But most of these models require retraining or finetuning, which may not be feasible in resource-constrained settings.

Motivated by the degradation of long-range dependencies in LVLMs and the limitations posed by resource constraints, we introduce an inference-only framework named \textbf{T}hree-stage \textbf{D}ecay-\textbf{R}esilient \textbf{S}trategies (T-DRS). Without model re-training, T-DRS collectively enhance attention robustness over extended token sequences through 1) \textbf{Semantic-Driven DRS (SD-DRS)}, 2) \textbf{Distance-aware Control DRS (DC-DRS)}, and 3) \textbf{re-Reinforce Distant DRS (reRD-DRS)}. In Fig.~\ref{intro}(B.1), the SD-DRS initially introduces a content-aware residual term into the pre-softmax attention logits to strengthen semantically aligned but distant token interactions, ensuring that vital contextual information is not prematurely suppressed. Having the potential of perturbed local structures, DC-DRS in Fig.~\ref{intro}(B.2) enforces a locality-aware prior, yielding a more structured and focused attention distribution, with its curved form in Fig.~\ref{intro}(C) functions as a smoothing and distance-aware mechanism that reduces attention weights. To further cope with the tokens that are distant but rich in semantics, we have reRD-DRS in Fig.~\ref{intro}(B.3). It introduces a reinforcement mechanism that selectively restores attention mass to under-attended yet semantically important long-range pairs, compensating for cumulative decay effects while preserving the local dependencies established by the previous DRS. Together, the T-DRS offers a principled and interpretable solution for preserving both global coherence and local precision in long-context reasoning. The entire pipeline operates at inference time, remains fully differentiable, and is compatible with existing transformer architectures.

\paragraph{Our main contributions are summarized as follows:}
\begin{itemize}
\item We propose \textbf{T-DRS}, a training-free, inference-only framework that alleviates long-range attention decay through three complementary decay-resilient strategies.
\item \textbf{Distant recovery and locality smoothing}: SD-DRS strengthens distant semantic links via content-aware residuals, while DC-DRS imposes a smooth locality bias to refine attention allocation.
\item \textbf{Residual re-weighting}: reRD-DRS reinforces semantically important long-range dependencies by recovering suppressed attention weights, thereby enhancing global reasoning capabilities.
\item Extensive experiments on benchmark vision-language answering tasks demonstrate that T-DRS consistently outperforms state-of-the-art baselines, especially under long-context and reasoning-heavy scenarios.
\end{itemize}

\section{Related Work}
\subsection{Multimodal Learning with LVLMs}
LVLMs have become a dominant paradigm for unified multimodal understanding and generation~\cite{xue2025mmrc}. By integrating high-capacity vision encoders with large-scale language models, LVLMs are capable of processing interleaved image and text inputs and generating free-form responses~\cite{radford2021learning, touvron2023llama, fang2023eva, tangunivit}. Recent advances leverage instruction tuning~\cite{ouyang2022training, zhang2022opt} to align vision and language modalities under unified prompts. Models such as Flamingo~\cite{alayrac2022flamingo}, BLIP-2~\cite{li2023blip}, and MiniGPT-4~\cite{zhu2023minigpt} introduce lightweight adapters (i.e., Q-Formers) to efficiently inject vision into language decoders. LLaVA~\cite{liu2023visual, an2025llava} integrates CLIP visual features directly into a Vicuna-style decoder, demonstrating impressive performance on a wide range of multimodal benchmarks without explicit cross-attention layers. These models have significantly advanced downstream tasks such as visual question answering, image-text retrieval, and multimodal dialogue, paving the way for unified multimodal agents.

\subsection{Position Embeddings in LVLMs}
PE plays a crucial role in LVLMs, as it informs the model of token order and structural relationships across modalities. Early vision-language models typically adopt absolute~\cite{dosovitskiy2020image} or learnable~\cite{li2021align} positional embeddings, often applied independently to text and image tokens. While effective in short sequences, these methods generalize poorly to long or variable-length contexts. To address this, recent LVLMs adopt RoPE~\cite{su2024roformer}, a relative encoding mechanism that applies sinusoidal rotations to query and key vectors. RoPE encodes distance implicitly, supports sequence length extrapolation, and avoids additional parameters, making it especially suitable for decoder-only architectures. However, RoPE inherently introduces a long-range decay effect: as the relative distance between tokens increases, their attention scores diminish due to the orthogonal nature of high-angle rotations.

While such decay aligns with the local inductive bias in language modeling, it can be problematic in multimodal tasks like VQA~\cite{lu2022learn, agrawal2018don,zhang1,zhang2,zhang3,zhang4,zhang5,zhang6,zhang7}, where image and text tokens are often separated by large positional gaps. For example, a question word appearing early in the sequence may need to attend to a relevant visual region encoded much later. In these cases, RoPE’s distance-based suppression may weaken essential cross-modal interactions, impairing the model’s ability to align semantics across modalities. In this work, we propose a semantic-aware strategy to enhance RoPE's long-range attention capabilities in multimodal contexts.

\section{Preliminary}
In this section, we present pre-definitions of LVLMs, the baseline method RoPE, and our motivation to have T-DRS.

\subsection{Large Vision-Language Models}
Large Vision-Language Models (LVLMs). Given a pretrained vision encoder ${F}_{v}$ and a lightweight projection head $f$, the visual content ${I}_{v}$ is projected into the embedding space of the Large Language Model (LLM) as follows:
\begin{equation}
     S_{\text{vision}}
  = f\bigl({F}_{v}({I}_{v})\bigr)
  = \{w^v_0, w^v_1, \dots, w^v_{V-1}\}.
\end{equation}
For the instruction prompt ${I}_{t}$, the LLM encodes it into a $T$ textual tokens language embedding as
\begin{equation}
    S_{\text{instr}}
  = F_t(I_t)
  = \{w^t_0, w^t_1, \dots, w^t_{T-1}\}.
\end{equation}
$S_\text{vision}$ and $S_\text{instr}$ are concatenated.
\begin{equation}
  S = \bigl\{S_{\text{vision}},\,S_{instr}\bigr\}
    \;\in\;\mathbf{R}^{(V+T)\times d},
    \label{3}
\end{equation}
into a single multimodal token sequence, where $d$ is the shared embedding dimension. The cross-modal attention layers interleave and attend to all the $V+T$ tokens, seamlessly fusing visual and textual information to facilitate downstream generation.

\subsection{Rotary Positional Embedding}
Rotary Positional Embedding (RoPE) encodes the position of the token through a rotation matrix applied to each embedded token. For a token $w_m$ at position $m \in [1, V+T]$, the corresponding rotation matrix
\begin{equation}
\resizebox{0.98\linewidth}{!}{$
R_{\boldsymbol{\theta}, m}^d = \mathrm{diag} \left(
\begin{bmatrix}
\cos m\theta_1 & -\sin m\theta_1 \\
\sin m\theta_1 & \cos m\theta_1
\end{bmatrix},
\ldots,
\begin{bmatrix}
\cos m\theta_{d/2} & -\sin m\theta_{d/2} \\
\sin m\theta_{d/2} & \cos m\theta_{d/2}
\end{bmatrix}
\right)
$}
\label{eq:rope_matrix}
\end{equation}
is constructed as a block diagonal matrix composed of $\frac{d}{2}$ two-dimensional rotations, where the rotation frequencies $\{\theta_i\}_{i=1}^{d/2}$ follow a predefined sinusoidal schedule $\theta_i = 10000^{-2(i-1)/d}$. In practice, RoPE is applied to both query and key vectors across all layers of Transformer-based LVLMs. This design encodes the relative distance between tokens directly into the attention score. The attention logits $A$ between query $Q_i$ and key $K_j$ are computed as:
\begin{equation}
A_{i,j} = \mathrm{softmax} \left( \frac{Q_i^\top R_{j-i} K_j}{\sqrt{d}} \right),
\label{eq:rope_attention}
\end{equation}
where $R_{j-i} = (R_{\boldsymbol{\theta}, i}^d)^\top R_{\boldsymbol{\theta}, j}^d$ is the relative rotation matrix based on the positional gap $j - i$. 

\subsection{Motivation of T-DRS}
While RoPE provides an elegant and efficient way to encode relative positions, its inherent design leads to a phenomenon known as \textit{long-range decay}~\cite{wei2025videorope}. As the relative distance $|j - i|$ increases, the effective attention score $A_{i,j}$ between tokens $Q_i$ and $K_j$ decreases due to the rotational orthogonality introduced by large-angle embeddings. This behavior aligns well with language modeling tasks, where distant tokens are typically less semantically related. However, in multimodal question answering tasks, the distant dependencies are less important often does not hold. In such settings, a question token may need to attend to visual features located far apart in the sequence, especially when the input consists of concatenated long textual instructions and high-dimensional visual tokens. RoPE's tendency to suppress long-distance attention can thus hinder effective cross-modal alignment, as relevant visual evidence may be located dozens or even hundreds of tokens away from the query. This results in a mismatch between semantic importance and attention strength.

Motivated by this limitation, we aim to mitigate RoPE’s long-range decay, allowing the model to better preserve meaningful dependencies across long token spans. To this end, we introduce T-DRS, a simple yet effective strategy designed to reinforce long-range attention.

\section{The proposed T-DRS Framework}

\begin{figure*}[t]
    \centering
    \includegraphics[width=\linewidth]{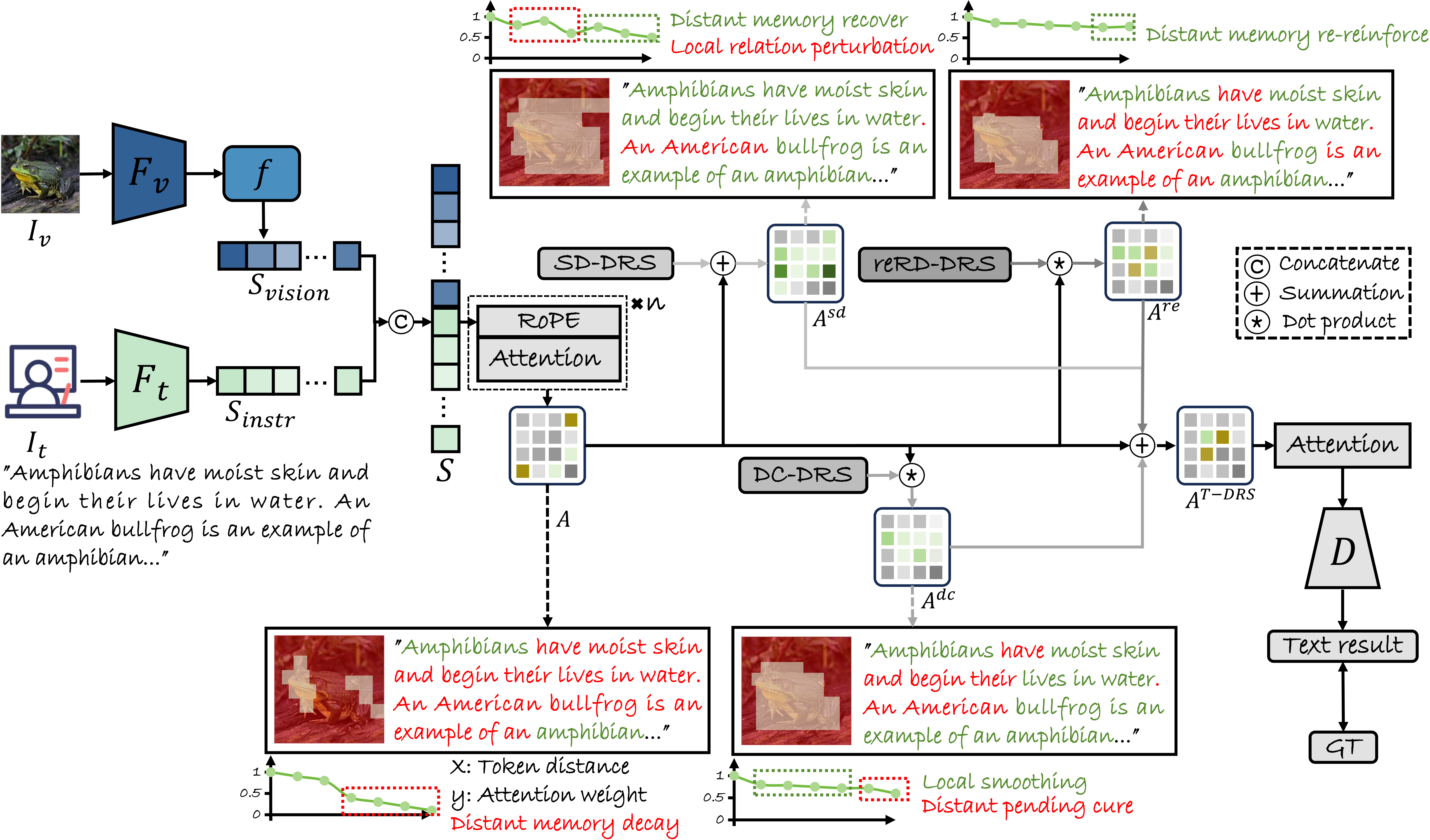}
    \caption{The proposed T-DRS is an inference-only framework. Given image and text inputs, visual and instruction features ($\textit{S}_{vision}$, $\textit{S}_{instr}$) can be extracted, then concatenated as $S$, input into the RoPE-attention architecture. The output attention logits $A$ experienced long-range dependencies decay, we then modulated three DRS: (1) SD-DRS ($A^{sd}$) amplifies semantically relevant distant tokens; (2) DC-DRS ($A^{dc}$) applies local smoothing technique, avoiding the local structure being perturbing, and (3) reRD-DRS ($A^{re}$)  is sepecialized for re-strengthens long-range attention with rich semantics. Integrating the pre-softmax attention map, $A^{T-DRS}$ is used to generate the final output.}
    \label{mainframe}
\end{figure*}

This section provides a detailed description of the T-DRS framework, containing three consecutive strategies: 1) Semantic‑Driven DRS (SD-DRS), 2) Distance-aware Control DRS (DC-DRS), and 3) re-Reinforce Distant DRS (reRD-DRS). The framework is shown in Fig. \ref{mainframe}. 

\subsection{Semantic‑Driven DRS (SD-DRS)}
Firstly, to address the visual regions and relevant text descriptions that reside at distant positions but are highly correlated, we have SD-DRS for recovery, which introduces a lightweight, training-free semantic bias to enhance $A$ from Eq.(\ref{eq:rope_attention}). Under the influence of RoPE, the standard softmax attention no longer treats all token pairs equally. Although it assumes that the dot product sufficiently captures relevance, this assumption becomes invalid for distant yet semantically important pairs, leading to unfair suppression of long-range dependencies. To mitigate this, we explicitly model semantic correlation beyond positional proximity. The key intuition is that semantically related token pairs should retain high attention weights. To this end, SD-DRS first computes a semantic affinity map:
\begin{equation}
{\text{sem}\_\text{sim}}_{i,j} = \cos(Q_i, K_j) = \frac{Q_i \cdot K_j}{\|Q_i\| \cdot \|K_j\|},
\end{equation}
which captures the cosine similarity between query and key embeddings. To make the values compatible with attention logits, we normalize the similarity to a positive range:
\begin{equation}
{\text{sem}\_\text{pos}}_{i,j} = \frac{1}{2} \left( {\text{sem}\_\text{sim}}_{i,j} + 1 \right),
\end{equation}
yielding a matrix values in \([0,1]\), higher values indicate stronger semantic correspondence. In the final output of the SD-DRS, we add the bias to the original logits as:
\begin{equation}
A^{sd}_{i,j} = A_{i,j} + {\text{sem}\_\text{pos}}_{i,j},
\end{equation}
effectively amplifying attention between semantically meaningful token pairs, especially those that are distant in position. This introduces a content-aware bias that complements the position-centric nature of traditional transformers, thus restoring the model's ability to capture long-range semantic dependencies. To modulate attention in a stable and bounded manner, we transform the semantic position similarity map ${\text{sem}\_\text{pos}}_{i,j}$ into a continuous scaling factor
\begin{equation}
\text{scale}_{i,j} = \frac{{\text{sem}\_\text{pos}}_{i,j} - \min(\text{sem}\_\text{pos})}{\max(\text{sem}\_\text{pos}) - \min(\text{sem}\_\text{pos})},
\label{scale}
\end{equation}
instead of directly injecting a semantic bias into the attention logits, which may destabilize training or inference. This normalization rescales each semantic position similarity score into the $[0, 1]$ range, enabling adaptive and bounded control over the two subsequent attention decay strategies.

\subsection{Distance-aware Control DRS (DC-DRS)}
While semantic cues are vital for guiding attention, local structure remains an essential inductive bias in sequential modeling. In Fig. \ref{mainframe}, although SD-DRS recovers distant token pairs, it slightly perturbs the local structure. To address this, we propose DC-DRS, a structure-preserving attention modulation smoothing method that explicitly incorporates token separation effects into the attention calculation.

DC-DRS formulates a smooth, distance-dependent attenuation profile with explicit analytic guarantees. Concretely, for a query at position $i$ and a key vector at position $j$, we define the relative distance as $d_{i,j} = |i - j| $, and enforce the following design criteria on the decay function $w(d)$ inspired by ~\cite{bishop2006prml}:
\begin{itemize}
    \item \textbf{Monotonicity:} $w(d)$ must strictly decrease with $d$, respecting the intuition of locality bias.
    \item \textbf{Smoothness:} To maintain compatibility with gradient-based optimization, the decay profile is required to be continuous and differentiable.
    \item \textbf{Lower-bound preservation:} A non-zero minimum attention value $w_{\min}^{dc}$ should be enforced at the maximum distance $d_{\max}$, to have persistent connectivity throughout the entire sequence.
\end{itemize}

To satisfy these constraints, we construct a closed-form attenuation profile parameterized by a decay scale $\sigma_0$, calibrated such that:
\begin{align}
\small
w(d_{i,j}) &= \exp\left(-\frac{1}{2} \left( \frac{d_{i,j}}{\sigma_0} \right)^2 \right), \nonumber \\ 
\small
\text{where }\sigma_0 &= \frac{\max(d_{i,j})}{\sqrt{-2 \ln w_{\min}^{dc}}},
\end{align}
where \( d_{i,j} \) denotes the positional distance between tokens at positions $i$ and $j$. The hyperparameter \( w_{\min}^{dc} \) specifies the minimum allowable attenuation weight at the maximal distance. This formulation ensures rapid decay for nearby positions while retaining a controlled residual weight for distant ones, embodying an interpretable, differentiable, and computationally efficient geometric prior. To adaptively modulate this decay according to semantic context, we incorporate scaling factors from Eq. (\ref{scale}), defining an effective distance:
\begin{equation}
\hat{d}_{i,j} = \frac{d_{i,j}}{\text{scale}_{i,j}}.
\end{equation}
With a high semantic alignment, the scaling factor can increase, effectively reducing \( \hat{d}_{i,j} \), softening the decay and allowing semantically important distant token pairs to maintain stronger attention. The distance-aware modulation weight is formulated as:
\begin{equation}
\small
r_{i,j}^{dc} = \exp\left( -\frac{1}{2} \left( \frac{\hat{d}_{i,j}}{\sigma_0} \right)^2 \right),
\end{equation}
which is multiplied by the $A$, returning
\begin{equation}
A_{i,j}^{dc} = \lambda_{dc} \cdot A_{i,j} \cdot r_{i,j}^{dc}
\label{13}
\end{equation}
as the final result.

Generally, DC-DRS aims to achieve a delicate balance, which enforces an inductive prior that encodes a smooth and bounded notion of locality, modulated by contextual relevance, and the local contextual information retains the long-range connectivity recovered by SD-DRS.

\subsection{re-Reinforce Distant DRS (reRD-DRS)}

Despite the dual guidance of semantic and positional priors, some semantically important but distant token pairs still receive weak attention due to compounded decay effects, as shown in the two final points of the last curve in Fig.~\ref{mainframe}. Then we have the final component, reRD-DRS, to further reinforce. It has a target modulation term that is specialized for re-weighting extreme cases, where token pairs with extremely high semantic affinity and significant distance.

We first define a re-weighting coefficient \( r_{i,j}^{re} \) as a reinforcement gate by a rational quadratic function which serves as a heavy-tailed kernel. The rational quadratic function~\cite{rasmussen2006gaussian} decays slower, allowing stronger reinforcement for long-range dependencies. :
\begin{equation}
r_{i,j}^{re} = \left(1 + \frac{(d_{i,j})^2}{2\cdot (\sigma_{re} \cdot \text{scale}_{i,j})^2} \right)^{-\alpha},
\end{equation}
Specifically, we compute:
\begin{equation}
\alpha = \frac{-\log(w_{\min}^{re})}{\log\left(1 + \frac{d_{\max}^2}{2 \cdot (\sigma_{re} \cdot \text{scale}_{i,j})^2} \right)}.
\end{equation}
This formulation guarantees that the reinforced attention maintains a lower bound at long range, while remaining sensitive to semantic similarity and contextual scales. It also eliminates the need for manually tuning the decay sharpness, making the model more robust and interpretable. This constraint is formalized as:
\begin{equation}
r_{i,j}^{re} \Big|_{d_{i,j} = d_{\max},\, \text{scale}_{i,j} = 1} = w_{\min}^{re},
\end{equation}
which yields the closed-form expression for  \( \sigma_{re} \):
\begin{equation}
\sigma_{re} = \frac{d_{\max}}{\sqrt{2\alpha(w_{\min}^{re (-1/\alpha)} - 1)}}.
\end{equation}
This ensures attention to the most distant tokens is smoothly decayed to exactly \(w_{\min}^{re}\), while nearby tokens retain stronger reinforcement. The final attention logit of reRD-DRS is 
\begin{equation}
A_{i,j}^{re} = \lambda_{re} \cdot A_{i,j} \cdot r_{i,j}^{re}.
\end{equation}
\( \lambda_{re} \) is a tunable coefficient controlling the reinforcement strength. This residual formulation ensures that no token pair is prematurely discarded due to positional distance alone, as long as its semantic relevance warrants attention. It introduces a controlled, content-aware reinforcement path for long-range interactions, without disrupting the local attention patterns established earlier. The attenuation profile in reRD-DRS decays smoothly with distance while maintaining a heavy-tail behavior, allowing attention to persist beyond the short-range regime. 

The final attention logits of T-DRS are modulated through a residual combination of all three DRS:
\begin{equation}
A_{i,j}^{T-DRS} = A_{i,j} + A_{i,j}^{sd} +A_{i,j}^{dc} + A_{i,j}^{re}.
\end{equation}
In essence, T-DRS integrates semantic cues and distance-aware priors in a three-stage pipeline to robustly preserve long-range attention in vision-language models.

\section{Experiment}
\subsection{Experimental Settings}
\textbf{Datasets.} We adopt three standard benchmark datasets: ScienceQA-IMG~\cite{lu2022learn}, GQA~\cite{hudson2019gqa}, and TextVQA~\cite{ganz2023towards} to evaluate model performance across diverse vision-language reasoning tasks. To further assess the effectiveness of our proposed T-DRS strategy in alleviating long-range dependency degradation, we additionally employ the Positional Object hallucination Prevalence Evaluation (POPE) dataset~\cite{li2023evaluating}, which specifically targets positional hallucination phenomena. ScienceQA-IMG is a curriculum-based benchmark with 21,208 multimodal questions requiring integration of visual and scientific knowledge. GQA offers 22M structured reasoning questions grounded in visual genome images; we use its balanced version for fair evaluation. TextVQA emphasizes OCR-based reasoning over scene text with 28K Q\&A pairs from OpenImages. POPE tests factual grounding and spatial awareness by detecting hallucinations in object reasoning. These datasets comprehensively evaluate structural reasoning, and cross-modal understanding. We report standard accuracy for all datasets, with the F1-score added for the POPE dataset.

\textbf{Comparison Methods.} Fourteen LVLM approaches i.e, Instruct-BLIP-7B \& 13B~\cite{dai2023instructblip}, BLIP-2-13B~\cite{li2023blip}, Shikra~\cite{chen2023shikra}, GPT3.5~\cite{zheng2023ddcot}, Ying-VLM~\cite{li2023m}, MiniGPT-4~\cite{zhu2023minigpt}, Qwen-VL-Chat~\cite{bai2023qwen}, Qwen-VL~\cite{bai2023qwen}, MobileVLM-v2-7B~\cite{chu2024mobilevlm}, Otter~\cite{li2025otter}, LLaVA1.5-7B~\cite{liu2023visual}, InterVL2-8B~\cite{chen2024far}, and Qwen2.5-VL-7B~\cite{bai2025qwen2} are chosen for comparison.

\textbf{Backbone Models.} To demonstrate the applicability of our method, we plug-in T-DRS into three representative LVLMs: LLaVA1.5-7B, Inter2VL2-8B, and Qwen2.5-VL-7B. These models vary in architecture and multimodal fusion strategies, allowing us to test the robustness and compatibility of T-DRS across various transformer designs.

\textbf{Implementation Details.} All models are tested in their original settings without further fine-tuning. T-DRS modules are injected at inference time, require no additional training, and are fully parameter-free except for a small number of fixed hyperparameters (\( w_{\min}^{dc},w_{\min}^{re}, \lambda_{\mathrm{dc}}, \lambda_{\mathrm{re}} \)), which are shared across all experiments for consistency. All the experiments are done on one NVIDIA A100 GPU.

\begin{table}[t]
\centering

\scalebox{0.83}{%
\label{tab1}
\begin{tabular}{c|ccc|cc}
\toprule
\multirow{2}{*}{Method} & \multicolumn{3}{c|}{QA Datasets} & \multicolumn{2}{c}{POPE} \\ \cmidrule{2-6} 
                        & Sci. & GQA & TextVQA & Acc. & F1-score \\ \midrule
Instruct-BLIP-7B    & 60.0 & 49.2 & 60.5 & 70.1 & 72.3 \\
Instruct-BLIP-13B   & 63.1 & 49.5 & 63.1 & 71.0 & 73.0 \\
BLIP-2-13B                   & 61.3 & 36.4 & 42.5 & 65.3 & 67.2 \\
Shikra                  & 45.8 & -    & -    & 63.2 & 65.8 \\
GPT3.5                  & 72.5 & -    & -    & 74.6 & 76.5 \\
Ying-VLM                       & 55.7 & -    & -    & 69.4 & 71.0 \\

MiniGPT-4               & 42.3 & 32.2 & -    & 59.7 & 62.1 \\
Qwen-VL-Chat               & 68.2 & -    & -    & 75.1 & 76.9 \\
Qwen-VL                   & 67.1 & -    & -    & 76.2 & 77.1 \\
MobileVLM-v2-7B       & 61.0 & 62.6 & -    & 78.0 & 78.5 \\
Otter                      & 66.3 & -    & -    & 71.9 & 73.7 \\
LLaVA1.5-7B      & \underline{67.9} & \underline{62.0} & 58.2 & \underline{83.3} & \underline{85.7} \\
InterVL2-8B        & \underline{96.6} & \underline{62.6} & 79.1 & \textbf{88.0} & \underline{87.0} \\
Qwen2.5-VL-7B       & \underline{79.4} & \underline{57.9} & 84.5 & \underline{87.7} & \underline{86.4} \\ \midrule
LLaVA1.5. + T-DRS                     & \textbf{69.2} & \textbf{63.1} & \textbf{59.0} & \textbf{83.7} & \textbf{86.1} \\
InterVL2. + T-DRS                   & \textbf{97.3} & \textbf{62.8} & \textbf{79.7} & \textbf{88.0} & \textbf{87.4} \\
Qwen2.5. + T-DRS                   & \textbf{80.7} & \textbf{58.3} & \textbf{85.0} & \textbf{88.5} & \textbf{87.3} \\
\bottomrule
\end{tabular}
}
\caption{Accuracy (Acc.\%) performance comparison on the ScienceQA-IMG (Sci.), GQA, TextVQA, and POPE datasets. All models operate in a training-free setting. T-DRS is plugged into three distinct LVLMs for evaluation.}
\raggedright
\footnotesize
\textit{note:} Best and second-best results are \textbf{bolded} and \underline{underlined}.
\end{table}

\subsection{Quantitative Comparisons}
To evaluate the generality of T-DRS, we integrate it into three diverse LVLMs, differing in scale and architecture. Without any fine-tuning, T-DRS consistently improves performance across ScienceQA, GQA, and TextVQA in Tab.~\ref{tab1}, with the improvement of 1.3\%, 1.1\%, and 0.8\% integrating with LLaVA, 0.7\%, 0.2\%, and 0.6\% with InterVL, and 1.3\%, 0.4\%, and 0.5\% with Qwen. demonstrating its plug-and-play flexibility and highlighting long-range attention degradation as a common limitation in current VLMs. Moreover, we evaluate our model on the POPE benchmark, which is specifically designed to expose hallucination errors that arise when models fail to ground their predictions in the actual visual content. The additional analysis provides stronger empirical evidence for the robustness and reliability of T-DRS across diverse reasoning scenarios. To further validate the effectiveness of our proposed method, we conduct comprehensive ablation studies to disentangle the contributions of each component within T-DRS.

\begin{table}[t]
\centering
\scalebox{0.8}{
\label{ablation}
\begin{tabular}{cc|ccc}
\toprule
\multicolumn{2}{c|}{\multirow{3}{*}{Configuration}} & \multicolumn{3}{c}{Datasets}        \\ \cmidrule{3-5} 
\multicolumn{2}{c|}{}                               & Sci.     & \multicolumn{2}{c}{POPE} \\
\multicolumn{2}{c|}{}                               & Acc.     & Acc.         & F1-score        \\ \midrule
\multirow{3}{*}{Baseline}           & LLaVA1.5-7B        & 67.9        & 83.3            & 85.7         \\
                                    & InterVL2-8B      & 96.6        & \underline{88.0}            & 87.0         \\
                                    & Qwen2.5-VL-7B         & 79.4        & 87.7            & 86.4         \\ \midrule
\multirow{3}{*}{\textit{w.}SD-DRS}           & LLaVA1.5-7B        & 68.1        & 83.4            & 85.8        \\
                                    & InterVL2-8B      & 96.9       & 87.6            & 86.8         \\
                                    & Qwen2.5-VL-7B         & 79.8        & 87.8            & 86.6         \\ \midrule
\multirow{3}{*}{\textit{w/o.}reRD-DRS}       & LLaVA1.5-7B        & \underline{68.8}        & \underline{83.6}            & \underline{86.0}         \\
                                    & InterVL2-8B      & \underline{97.1}       & 87.9            & \underline{87.2}        \\
                                    & Qwen2.5-VL-7B         & \underline{80.4}        & \underline{88.0}            & \underline{86.9}   \\ \midrule
\multirow{3}{*}{Full model}         & LLaVA1.5-7B        & \textbf{69.2}        & \textbf{83.7}            & \textbf{86.1}         \\
                                    & InterVL2-8B      & \textbf{97.3}        & \textbf{88.0}            & \textbf{87.4}       \\
                                    & Qwen2.5-VL-7B         & \textbf{80.7}        & \textbf{88.5}            & \textbf{87.3}            \\ \bottomrule
\end{tabular}
}
\caption{Ablation study of T-DRS components on ScienceQA-IMG (Sci.) and POPE, with each abatement item plugged in with LLaVA1.5-7B, InterVL2-8B, and Qwen2.5-VL-7B.}
\end{table}

\subsection{Ablation Studies}
To validate the contribution of each of the DRS, we conduct ablation studies in Tab. \ref{ablation}, utilizing the ScienceQA-IMG and POPE dataset, plugging in with the three distinct LVLMs. 

SD-DRS introduces semantic awareness into attention modulation, enhancing the model’s ability to attend to semantically aligned token pairs regardless of positional separation. However, this may occasionally perturb local structural dependencies, leading to performance fluctuations. To mitigate this, DC-DRS explicitly controls local structural coherence and achieves the second-best performance. Nonetheless, distant token pairs with semantically salient relevance still require enhanced attention recovery, which is addressed by reRD-DRS, specialized for residual long-range attention reinforcement. The integration of all three modules achieves the highest performance, underscoring their complementary roles and validating the effectiveness of the proposed decay-resilient framework.

\begin{figure}[t]
    \centering
    \includegraphics[width=1\linewidth]{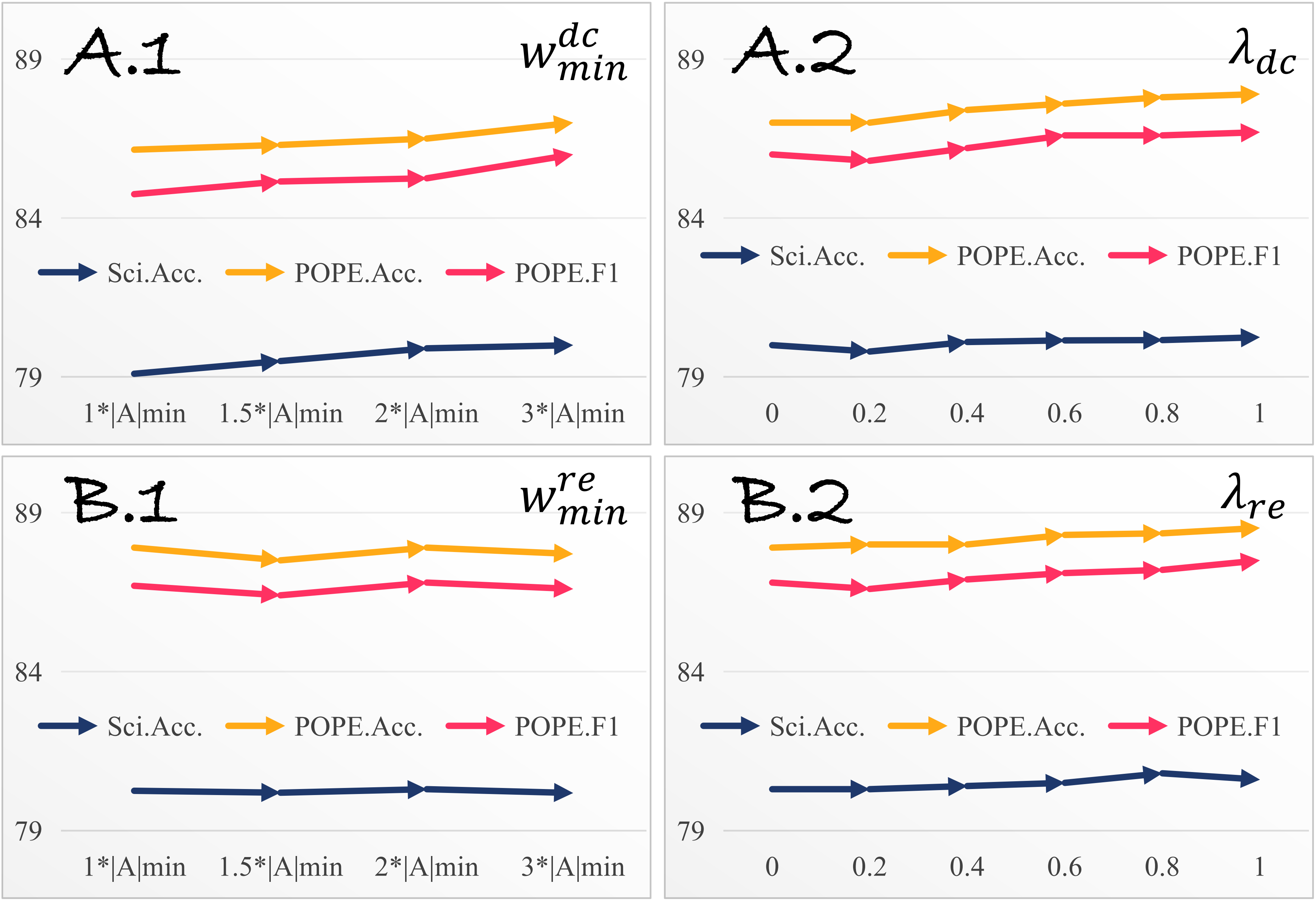}
    \caption{Four hyperparameters are evaluated using ScienceQA-IMG and POPE datasets. The values of $\lambda$ are determined following the selection of $w_{min}$, and the two hyperparameters in reRD-DRS are configured subsequent to the determination of those in DC-DRS. We assign the value of three times $\left|{A}\right|_{min}$ and 1 to the two hyperparameters in DC-DRS, and twice of $\left|{A}\right|_{min}$ with set 0.8 for $\lambda_{re}$ in ScienceQA-IMG, and 1 for POPE in reRD-DRS.}
    \label{hyper}
\end{figure}

\subsection{Qualitative Analysis}
Within the T-DRS framework, there are four hyperparameters \{$w^{dc}_{min}$, $\lambda_{dc}$, $w^{re}_{min}$, $\lambda_{re}$\} for sensitivity evaluation shown in Fig.~\ref{hyper}. The values of $\lambda$ are determined following the selection of $w_{min}$, and the two hyperparameters in reRD-DRS are configured subsequent to the determination of those in DC-DRS. $\left|{A}\right|_{min}$ denoted as the minimum value of the attention map within $A$ in Eq.(\ref{eq:rope_attention}).

$w^{dc}_{min}$ is the minimum attention weight of $A$ for distant tokens in the decay component, $\lambda_{dc}$ controls the strength of the distance-aware modulation weight $r_{dc}$ in Eq.(\ref{13}), $w^{re}_{min}$ specifies minimum reinforcement weight for distant but semantic relevance tokens, and $\lambda_{re}$, a coefficient to further decide the reinforcement strength of semantic relevance. When $w^{dc}_{min}$ is set to three times $\left|{A}\right|_{min}$ and $\lambda$ is fixed at 1, T-DRS (\textit{w/o.}reRD-DRS) configuration achieves the best performance in Fig.~\ref{hyper}(A). Upon further introducing reRD-DRS, setting $w^{re}_{min}$ to twice $\left|{A}\right|_{min}$ with $\lambda_{re}$ set to 0.8 for ScienceQA-IMG and 1 for POPE yields the highest overall performance for the full T-DRS model in Fig.~\ref{hyper}(B). Due to the role of $w_{min}$ in regulating attention to distant token pairs, a larger $w_{min}$ increases the model's tolerance for incorporating long-range interactions. However, excessive tolerance may cause semantically weak or irrelevant tokens to be attended, resulting in redundancy. Conversely, an overly small $w_{min}$ enforces overly strict filtering, potentially suppressing attention to nearby yet semantically meaningful tokens, thereby disrupting local structural coherence.

\begin{figure}[!htbp]
    \centering
    \includegraphics[width=1\linewidth, height=0.45\linewidth]{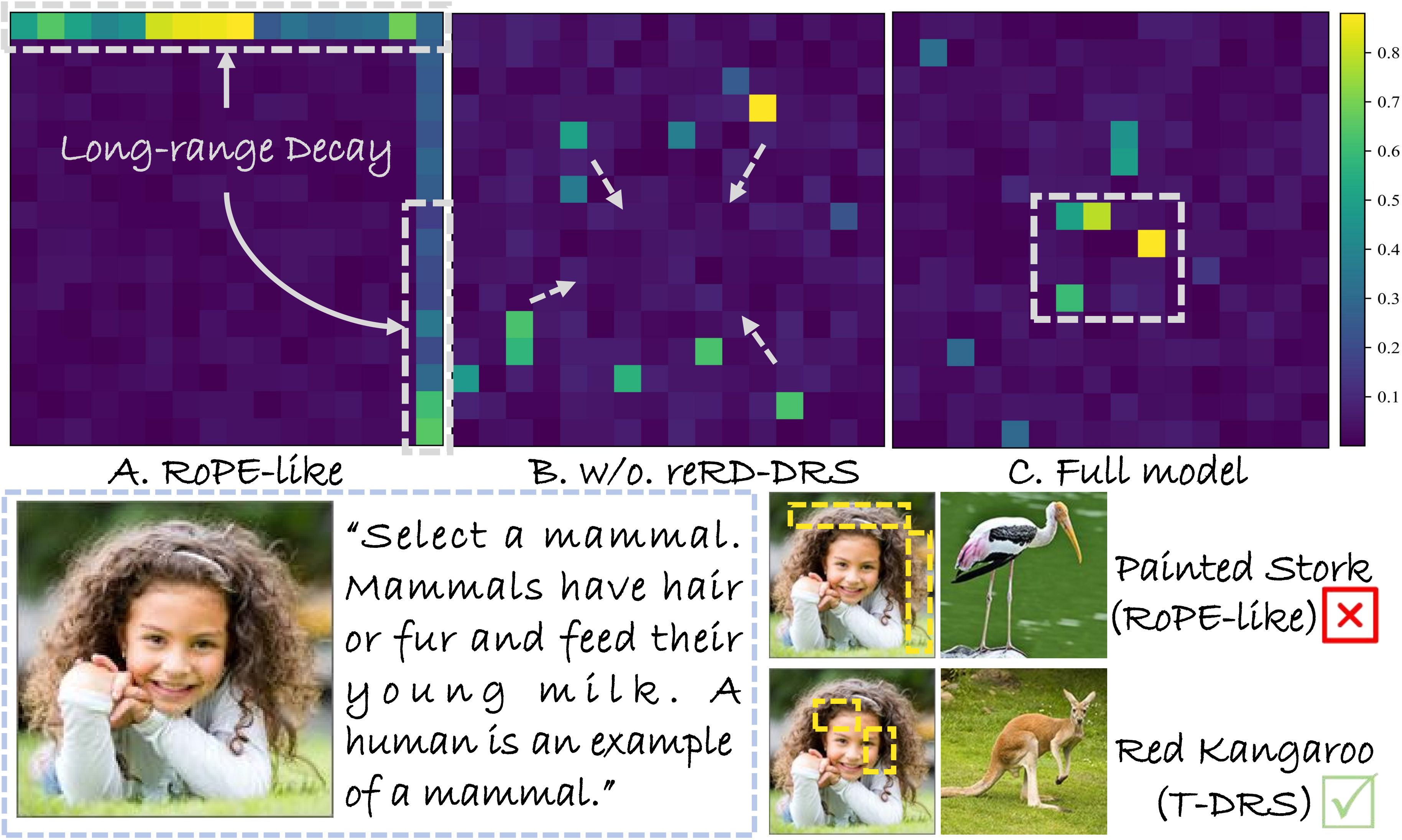}
    \caption{Visualization results of cross-attention of $S$ across different stages. (A) RoPE-only ($A$). (B) With SD-DRS and DC-DRS ($A+A^{sd}+A^{dc}$), attention begins to converge towards the central focus. (C) Under the full model setting ($A^{T-DRS}$), the model can further reinforce informative long-range connections.}
    \label{heatmap}
\end{figure}

\subsection{Visualization Results}
To better understand the joint effects of the distinct stages within T-DRS in the cross-modal reasoning process, we have the visualization results in  Fig.~\ref{heatmap}. In Fig.~\ref{heatmap}(A), RoPE-like approaches exhibit long-range decay, with tokens focusing primarily on image peripheries instead of semantically important regions. For instance, when presented with a mammal with furry hair, the model tends to focus on the fur, potentially leading to the incorrect selection of \textit{"Painted Stork"}, which has furry feathers. This strong locality bias risks missing key visual cues at the center of the image relevant to the task. Ultimately, the model may fail to capture the visual semantics needed for accurate cross-modal alignment. With the integration of SD-DRS and DC-DRS in Fig.~\ref{heatmap}(B), tokens shift noticeably toward the semantically important central regions of the image, and it shows a smooth activation pattern around the center, manifesting as a soft green glow, indicating consistent and focused attention. Under the full model setting in Fig.~\ref{heatmap}(C), after applying reRD-DRS, the attention becomes even more focused on semantically meaningful regions, reinforcing critical long-range token-to-token connections that are previously weak or under-attended. T-DRS can concentrate more on the object's face, correctly identifying it as a mammal and leading to the accurate selection of \textit{"Red Kangaroo"}.

\section{Conclusion}
In this paper, we propose \textbf{T}hree-stage \textbf{D}ecay-\textbf{R}esilient \textbf{S}trategies (T-DRS), an inference framework to alleviate long-range attetnions decay in Large Vision-Language Models (LVLMs). T-DRS integrates SD-DRS, recovering long-range dependencies, DC-DRS, smoothing local structure, and reRD-DRS specialized for reweighting tokens that are distant but with rich semantics. Experimental results demonstrate that T-DRS can be seamlessly applied to LVLMs.

\section{Acknowledgments}
The work is supported by Guangdong and Hong Kong Universities' “1+1+1” Joint Research Collaboration Scheme (2025A0505000003 ), the Guangdong Provincial Key Laboratory of Interdisciplinary Research and Application for Data Science, BNU-HKBU United International College (2022B1212010006), Guangdong Key Lab of AI and Multi-modal Data Processing (2020KSYS007), and the internal grant of BNBU (R6025A, UICR0300019).


\bibliography{aaai2026}

\end{document}